# Soft robotic hand with finger-bending/friction-reduction switching mechanism through 1-degree-of-freedom flow control

Toshihiro Nishimura[1], *Member, IEEE*, Kensuke Shimizu[2], Seita Nojiri[2], *Student member, IEEE*, Kenjiro Tadakuma[3], *Member, IEEE*, Yosuke Suzuki[1], *Member, IEEE*, Tokuo Tsuji[1], *Member, IEEE*, and Tetsuyou Watanabe[1], *Member, IEEE*

*Abstract*—This paper proposes a novel pneumatic soft robotic hand that incorporates a mechanism that can switch the airflow path using a single airflow control. The developed hand can control the finger motion and operate the surface friction variable mechanism. In the friction variable mechanism, a lubricant is injected onto the high-friction finger surface to reduce surface friction. To inject the lubrication using a positive-pressure airflow, the Venturi effect is applied. The design and evaluation of the airflow-path switching and friction variable mechanisms are described. Moreover, the entire design of a soft robotic hand equipped with these mechanisms is presented. The performance was validated through grasping, placing, and manipulation tests.

*Index Terms*—Grippers and Other End-Effectors, Hydraulic/Pneumatic Actuators, Soft Robot Materials and Design.

## I. INTRODUCTION

A novel system is proposed that can switch the flow path using a flow control in the development of a new pneumatic soft robotic hand realizing a surface friction change through a 1-degree-of-freedom (DOF) actuation with a single airflow. A soft robotic hand with deformable surfaces can stably grasp objects of various shapes owing to the large friction of the surface and the surface flexibility. These features have motivated the development of various soft robotic hands [1]–[3]. A pneumatic-driven system is the most popular for actuating a soft robotic hand. The finger of the hand is made of soft materials, such as silicone, and the air chambers are embedded in the deformable finger. Air pressure is applied to the chambers from the air source. The expansion of the chambers generates bending motions of the finger to activate the grasping motion. Finger bending motions of a pneumatic soft robotic hand can be generated by a single airflow through an air tube, allowing for a lightweight hand design [4]–[7].

A typical problem of a soft robotic hand, including pneumatic hands, is the difficulty in manipulating the slippage between a grasped object and the finger surface. As described previously, a large friction on the surface is effective for stably grasping an object. However, a stable or converged posture of the object resulting from the grasping motion could be unexpected because of the large friction and surface flexibility. One solution is an in-hand manipulation using slippage after grasping to control the object to its desired posture [8]. Unfortunately, the large friction and flexibility of the finger surface make it difficult to perform in-hand manipulation, which involves sliding, rolling, or transferring an object within the hand. Switching the surface friction between the high and low friction states is a solution to achieve in-hand manipulation. We previously proposed a variable friction surface mechanism that can operate on a flexible surface [9][10]. The mechanism consists of two structures: a slit-shaped surface texture generating high friction under both wet and dry conditions [9] and a friction-reduction system utilizing a lubricant (anhydrous ethanol) injection [10]. By adopting the mechanism for a soft robotic hand, the issue of the hand can be resolved. However, in [9][10], the lubricant injection and robotic hand operations were driven by different sources, such as electric motors for the robotic hand and a manual syringe operation for the lubrication. Integrating these systems with a simple control remains an open problem, and the installation of a lubrication mechanism for a multi-joint fingered robotic hand has not been realized.

To operate both the lubricant-injection system and pneumatic soft robotic hand motion control, two different actuation systems are required: lubricant (anhydrous ethanol) injection and pneumatic control systems. The use of different systems burdens the control of the robotic hands. Even if a pneumatic control system is used to control the injection, and a solenoid valve is applied to switch the control target to either injection or finger motion, control of the solenoid valve is still required. At least two control systems are necessary. The number of wires and air tubes mounted on a manipulator should also be minimized to enhance the reliability and durability of the entire manipulation system. To simplify the control schema, this study is aimed at developing a system for controlling both the

---

Manuscript received: October 11, 2021; Revised: January 8, 2022; Accepted February 15, 2022. This letter was recommended for publication by Associate Editor and Editor Cecilia Laschi upon evaluation of the reviewers' comments.

This work was supported by JST Moonshot R&D Grant Number JPMJMS2034, MEXT/JSPS KAKENHI Grant Number JP21H01286, JP21K19790, CAO/SIP "An intelligent knowledge processing infrastructure, integrating physical and virtual domains" (funding agency: NEDO).

[1]T. Nishimura, Y. Suzuki, T. Tsuji, and T. Watanabe are with the Faculty of Frontier Engineering, Institute of Science and Engineering, Kanazawa University, Kakuma-machi, Kanazawa city, Ishikawa, 9201192 Japan (e-mail: tnishimura@se.kanazawa-u.ac.jp, te-watanabe@ieee.org).

[2]K. Shimizu and S. Nojiri are with the Graduated school of Natural science and Technology, Kanazawa University, Kakuma, Kanazawa, 9201192 Japan.

[3]K. Tadakuma is with the Graduation School of Information Sciences, Tohoku University, Sendai, Japan

Digital Object Identifier (DOI): see top of this page.

injection and finger motions through a 1-DOF actuation system. A novel switching mechanism using airflow control was proposed for actuation. If the airflow reaches a certain value, the path of the airflow changes. One airflow path is used for finger motion control, and the other is used for injection. Thus, switching of the injection and finger motion control systems was achieved using airflow control alone. The finger motion was controlled by the airflow within a low airflow range. We applied the Venturi effect to inject ethanol with a high airflow. Only a single air source line installed on a manipulator operates both the robotic hand and the friction-reduction system.

The goals of our design are 1) a hand consisting of soft pneumatic fingers, 2) a texture creating a high surface friction and a Venturi-effect based lubricant injection mechanism installed on the fingers, and 3) finger motion control and lubricant injection applied through single airflow control utilizing the novel flow-path switching mechanism.

*A. Related Work*

Several pneumatic robotic hands have been proposed [1][11], driven by pressurization or depressurization of the fluid flowing in the body or fingers. A common finger bending method makes one side of the finger inextensible and the other side soft and inflatable. Deimel et al. proposed a soft five-fingered robotic hand with deformable and non-deformable finger parts bent through air pressurization [12]. In [6], the design of a hand including a 2-DOF finger with separated chambers and a manipulation strategy utilizing the developed hand were proposed. Zhou et al. proposed a 13-DOF soft robotic hand in which several air chambers are installed in each finger [13]. Ryan et al. designed a soft hand using three sensors [14]. Pressurization is also used to generate negative pressure for suction-based grasping. The vortex and Bernoulli grippers are representative suction grippers that utilize a mechanism for converting the input pressurization into depressurization [15]. Li et al. proposed a vortex gripper that can suck an object with a rough surface using a rotating airflow [16]. Dini et al. investigated the performance of Bernoulli gripper configurations for grasping leather products [17].

Several methods for changing the contact surface friction were also proposed. The contact area affects the surface friction. Some studies have focused on this feature for constructing a variable surface friction system. Suzuki et al. altered the size and shape of wrinkles on a silicone body using compression and decompression [18]. Liu et al. changed the size and shape of wrinkles by exposing them to UV light [19]. Abdi et al. used a capacitive microelectromechanical actuator to control the shape of the skin [20]. By contrast, Nojiri et al. developed a mechanism by which the contact surface area is passively controlled using a contact load to alter the contact friction [21] and developed a friction variable system by observing the contact area using a built-in camera [22]. A method by which a sticky surface is pushed out from holes or gaps can also increase the friction. Becker et al. developed a gripper with a built-in inflatable flexible balloon to extrude sticky prongs [23]. Spiers et al. developed a sticky surface extrusion mechanism that works passively with the contact load [24]. In addition, Lu et al. designed a gripper inspired by origami that can switch the surface friction [25]. Adhesion control can also be used to change the amount of friction. Kim et al. used a shape-memory polymer to adjust the adsorption force using temperature control [26]. Shintake et al. applied dielectric elastomer actuators to control the electroadhesion force for gripping objects [27]. Hawkes et al. utilized the unidirectional adhesion structure of geckos to construct a gripper with a shear grasping function [28], and Glick et al. improved the grasping performance [29]. A special mechanical structure can also change the friction. Golan et al. designed a mechanism for switching a friction state using a Swivel-based mechanism [30].

Several pneumatic mechanisms providing multiple functions using a single control system have been proposed. Tani et al. proposed a pneumatic soft actuator that generates a vibratory bending motion in two opposite directions using a single airflow by switching the airflow path according to the deformation of the actuator [31]. Preston et al. designed a soft oscillator driven by a single constant pressure and adopted it in a rolling robot that performs periodic motion [32]. Vasios et al. also proposed a four-legged walking soft robot operated by a single input airflow, utilizing the viscosity of the soft material [33]. Tsukagoshi et al. developed a soft actuator propelling in a thin pipe with a single pneumatic input [34]. Ben-Haim et al. also proposed a method for controlling a soft actuator with a single flow control using a viscous fluid [35], whereas Napp et al. designed a valve that switches the open/closed state according to the pressure conditions at the input and output ports of the valve [36]. The systems in [31]-[36] can also switch the pressurizing configuration with a single flow control; however, the configurations are limited to either binary (i.e., to pressurize or not pressurize the chambers) or constant periodical (e.g., oscillation). Fingers used in robotic hands should change their configuration or bending angles according to the size and weight of the target object. Neither binarized nor periodic motions can achieve such handling motions. By contrast, our system provides continuous variable configurations or postures of the fingers. In addition, the mechanisms in [35] and [36] require a longer time to achieve the target configuration because they use the viscosity of the fluid or change in the pressure condition.

Although not a pneumatic mechanism, Miyamoto et al. developed a mobile robot that can move forward and backward and turn using a single rotary actuator [37]. This mechanism is designed for rigid mobile robots and is difficult to apply to mechanisms for soft robot hands.

Although several pneumatic robotic hands and friction variable mechanisms have been proposed, no attempt has been made to integrate and control them independently through 1-DOF actuation. In addition, no mechanism for switching two completely different functions (finger motion and lubricant injection in this study) using a 1-DOF (single) actuation system has been developed.

## II. Design of developed finger and hand

### A. Functional requirement

The functional requirements of the developed soft robotic hand are 1) flexible fingers driven by pneumatic pressure, 2) embedding the variable surface-friction mechanism based on the lubricant injection in the fingers, and 3) both the finger motion and lubricant injection and the switching between them being controlled through a single airflow control.

### B. Overview of Developed Robotic Hand System

Fig. 1 shows a schematic of the developed hand system. Flow-channel-switching (FCS) and lubricant injection mechanisms are the keys to achieving switching between finger movement and friction change with 1-DOF airflow control. An FCS mechanism is mounted on the palm of the hand and can switch between the airflow paths for finger motion and lubricant injection according to the volumetric airflow rate controlled by the airflow controller. Note that this study focuses not on the air pressure, but on the amount of airflow with the volumetric airflow rate because the airflow pressure can change according to the cross-sectional area of the air path, irrespective of the constant volumetric airflow rate. If the airflow rate is low, airflow is supplied to the finger body for movement. If the airflow rate is high, the airflow is supplied to the lubricant injection mechanism. Using the Venturi effect, the lubricant injection mechanism draws lubricant (anhydrous ethanol) from the lubricant tank mounted on the palm and injects it into the finger surface from the output port at the middle of the finger to reduce the surface friction, [38]. By controlling the volumetric airflow rate to the FCS mechanism, the developed hand realizes finger-bending motion as well as changes in surface friction through 1-DOF actuation.

## III. Structure of Key Mechanisms

### A. FCS mechanism

This section describes the structure of the FCS mechanism, which can switch the airflow path according to the volumetric airflow rate. Fig. 2 shows a three-dimensional computer-aided design (3D-CAD) model of the prototype mechanism. Three air tubes (one input and two output tubes) are connected to the mechanism. The input tube (yellow part in Fig. 2) is supplied with airflow from the airflow controller (Kofloc, 8550MC). Output tube 1 (red part) is bifurcated from the input tube in the cover box (green parts) and passed outside the box to control the finger motion. Output tube 2 (purple part) is connected to the upper side of the cover box for lubricant injection. The main roles of the exhaust port are to release air inside the finger chambers when opening the finger, and to determine the airflow rate values to output tube 1 and tube 3 for switching. Note that the FCS mechanism requires a continuous airflow, even when the finger posture is not changed owing to port exhaustion. An L-shaped switching lever was installed in the cover box to switch the airflow path. The L-shaped tip is sufficiently sharp to block the flow of air in output tube 1. On the other side of the lever, a protrusion occurs. Air from the input tube is bifurcated into output tube 1 and tube 3. Air from tube 3 flows out of output tube 2 through the inside of the cover box. The protrusion blocks the airflow from tube 3 and prevents air from flowing out of output tube 2. Fig. 3 illustrates the behavior of the FCS mechanism when the input volumetric airflow rate is changed. If the flow rate is low (Fig. 3(a)), the lever does not rotate and the air route to output tube 2 is blocked; therefore, air flows out of output tube 1 only. This state is referred to as "State A." Fig. 3(b) illustrates the behavior when the input airflow rate is in the middle such that the lever rotates to open both the airflow paths to output tubes 1 and 2. This state is referred to as "State B." In this state, finger movement and lubricant injection can be achieved simultaneously. However, to ensure a wide range of finger motions, the lubricant injection system was designed to avoid activating the injection within this range of the flow rate (in State B). The design details are described in section III.B. Within this range, it is difficult to operate the finger motion control and lubricant injection independently. If the airflow rate increases and the lever is rotated to completely block the airflow of output tube 1, the finger motion stops. Because the air in the finger chamber cannot flow outside, the finger posture is maintained. By contrast, the output-tube-2 airflow activates the lubricant injection. This state is referred to as "State C" (see Fig. 3(c)). We developed a prototype of the FCS mechanism by tuning its dimensions through trial and error such that the previously described operations are available. The

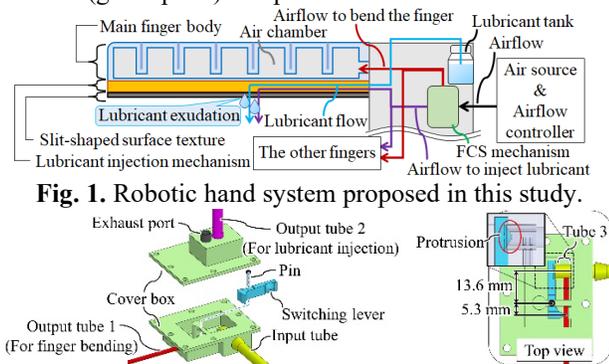

**Fig. 1.** Robotic hand system proposed in this study.

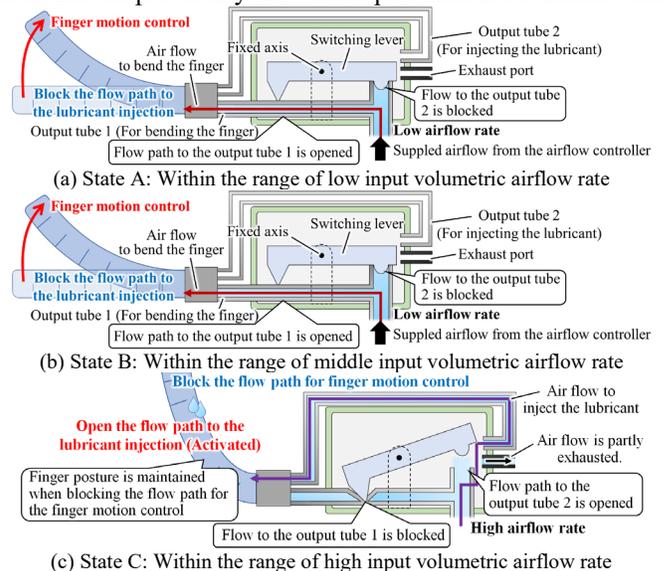

(a) State A: Within the range of low input volumetric airflow rate

(b) State B: Within the range of middle input volumetric airflow rate

(c) State C: Within the range of high input volumetric airflow rate

**Fig. 3.** Behavior of FCS mechanism when changing the airflow rate.

**Fig. 2.** 3D-CAD model of FCS mechanism.

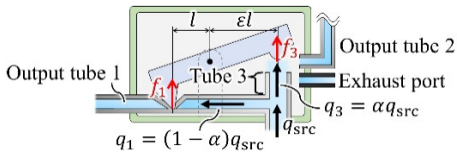

**Fig. 4.** Model of FCS mechanism

prototype is called "Prototype A."

Herein, we analyze the FCS mechanism to achieve the desired behavior of switching the airflow path by blocking output tube 1 according to the input airflow rate. The nomenclature used for the analysis is shown in Fig. 4. Let $q_1$, $q_3$, and $q_{src}$ be the volumetric airflow rates in output tube 1, tube 3, and the input tube, respectively. From the continuity equation, $q_1 = (1 - \alpha)q_{src}$ and $q_3 = \alpha q_{src}$. Here, $\alpha$ is the parameter that represents the ratio of airflow to output tube 1 and tube 3 and can be tuned by the size of the exhaust port: Here, $\alpha$ increases with an increase in the size of the exhaust port owing to the change in ratio in the cross-sectional area of the outputs. From the principle of conservation of momentum, the force, $f_3$, applied to the lever by the airflow from tube 3, is expressed by

$$f_3 = \rho q_3^2/s_3 = \rho \alpha^2 q_{src}^2/s_3, \quad (1)$$

where $\rho$ is the density of air, and $s_3$ is the cross-sectional area of tube 3. When the lever is rotated by $f_3$, the force $f_1$, applied to output tube 1 from its sharp tip, is given by

$$f_1 = \varepsilon f_3 = \varepsilon \rho q_3^2/s_3 = \varepsilon \rho \alpha^2 q_{src}^2/s_3, \quad (2)$$

where $\varepsilon$ is the ratio of the respective lengths from the pin of the lever to the acting points of $f_1$ and $f_3$. Let $f_{block}$ be the force required to block tube 1. Here, $f_1$ should satisfy $f_1 \geq f_{block}$ for blocking, and $f_{block}$ is determined based on the stiffness and cross-sectional shape of tube 1 and the air pressure inside the tube. The internal pressure was determined using $q_1$. Here, $q_1$ is changed by the amount of blockage in tube 1. Hence, the sufficient condition for blocking tube 1 is

$$f_1 = \varepsilon \rho q_3^2/s_3 = \varepsilon \rho \alpha^2 q_{src}^2/s_3 \geq f_{block}|_{q_1=q_1^{max}}, \quad (3)$$

where $q_1^{max}$ denotes the maximum $q_1$ during the blocking process. To obtain the desired behavior of the FCS mechanism, $\alpha$, $\varepsilon$, $s_3$, and the parameters affecting $f_{block}$ (i.e., the stiffness and cross-section of tube 1) should be tuned as the design parameters such that (3) is satisfied.

Next, we focused on the effects of $\alpha$ and $\varepsilon$ on the behavior of the FCS mechanism and experimentally examined them. First, the relationship between the airflow in output tube 1 $q_1$ and the $f_{block}$ was investigated. Fig. 5(a) shows the experiment setup. The force at which the value of the airflow meter became zero when a constant $q_1$ was given was measured as $f_{block}$. The results of $f_{block}$ when $q_1$ varies from 0 to 25 L/min are shown in Fig. 5(b). Subsequently, to validate the switching function of the FCS mechanism, the values of $q_1$ and $q_2$ were measured by varying the input $q_{src}$. The experiment setup is illustrated in Fig. 6. The input tube was connected to an air source through a regulator and an airflow meter. The outlets of the output tubes 1 and 2 were connected to the airflow meters. The input airflow rate was controlled by a regulator. The airflow rate of the input, $q_{src}$, was varied from 0 to 150 L/min, where 150 L/min is the maximum rate of the air source used. In addition to Prototype A, for comparison, we prepared "Prototype B" without an exhaust port, "Prototype C" whose $\varepsilon$ is small ($\varepsilon$ for Prototype

A = 2.6, and for Prototype B = 1.5), and "Prototype D" whose size of the exhaust port is large (the cross-sectional area of the port size for Prototype A is 7.1 mm², and for Prototype D is 50.3 mm²). Fig. 7 shows the transition of $q_1$, $q_2$, and $q_{src}$ when the input $q_{src}$ changed with the passage of time in the Prototype A test. At a low airflow rate (less than 8.1 L/min), the airflow flowed out of output tube 1 only (State A). At the middle airflow rate (8.1−118 L/min), the airflow flowed out of both output tubes 1 and 2 (State B). At a high airflow rate (over 118 L/min), the airflow flowed out of output tube 2 only (State C). The results demonstrate that the FCS mechanism in Prototype A can switch between the airflow paths to output tubes 1 and 2 by controlling the input volumetric airflow rate. Fig. 8 and Table I summarize the results of all prototypes. From the measured values of $q_1^{max}$ and $q_{src}^{max}$ shown in Table I, $f_1$ and $f_{block}$ were estimated (see Table I). In Prototypes A and D, (3) $f_1 \geq f_{block}$ was satisfied, whereas (3) was not satisfied in Prototypes B and C. In Prototype B, a small $\alpha$ induced a large $q_1^{max}$ and then a large $f_{block}$, whereas $f_1$ did not increase significantly. In Prototype C, $f_{block}$ was close to that in Prototype A, whereas a small $\varepsilon$ decreased $f_1$. In Prototype D, (3) was satisfied and the blocking of output tube 1 was achieved; however, $q_2$ was too small to activate the lubricant injection (see Fig. 8). The main reason for this is the large air loss from the exhaust port. Note that a large $q_1^{max}$ is preferable for generating a large tip force and bending deformation at the soft fingers. For a large $q_1^{max}$, $\alpha$ should be small, whereas blocking becomes difficult. Therefore, tuning the size of the exhaust port, namely, $\alpha$, is important. Practically speaking, $\alpha$,

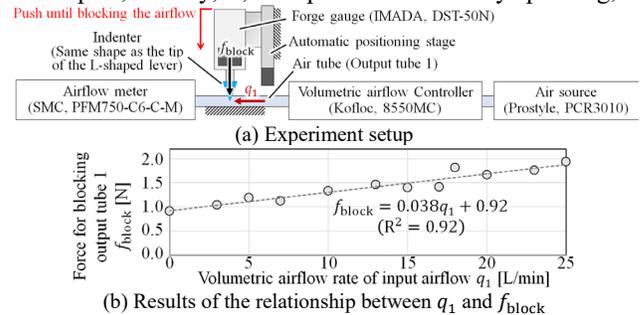

**Fig. 5.** Relationship between the airflow, $q_1$, and $f_{block}$

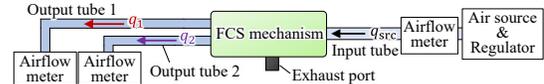

**Fig. 6.** Experimental setup used to measure $q_1$, $q_2$, and $q_{src}$

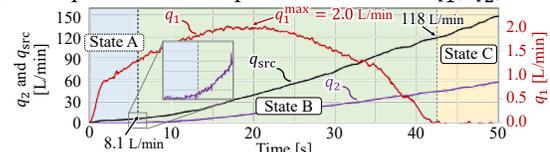

**Fig. 7.** Results of airflow rates from output tubes 1 and 2 and input airflow rate from the air source in Prototype A

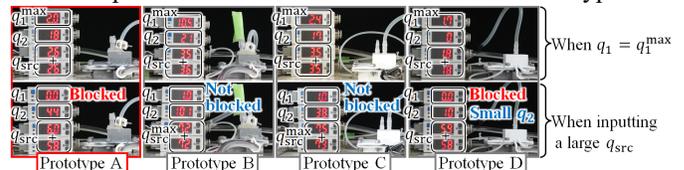

**Fig. 8.** Results of volumetric airflow rates for each prototype.

TABLE I  MEASURED AIRFLOW RATE AND ESTIMATED FORCE

| Proto-type | Designed | | Measured | | | Estimated | | |
|---|---|---|---|---|---|---|---|---|
| | Exhaust port size [mm²] | $\varepsilon$ [-] | $q_1^{max}$ [L/min] | $q_{src}^{max*1}$ [L/min] | Blocking tube 1 | $q_3^{*2}$ [L/min] | $f_1^{*3}$ [N] | $f_{block}^{*4}$ [N] |
| A | 7.1 | 2.6 | 2.0 | 118 | Success | 116 | 1.01 | 0.99 |
| B | 0 (None) | 2.6 | 10.5 | 144 | Failure | 134 | 1.18 | 1.34 |
| C | 7.1 | 1.5 | 2.4 | 148 | Failure | 146 | 0.90 | 1.02 |
| D | 50.3 | 2.6 | 1.7 | 117 | Success | 115 | 0.99 | 0.98 |

*1 In Prototypes A and D, this value is the value when state B changes to C. In Prototypes B and C, these values were measured when the air compressor was set to the maximum output.
*2 Estimated by the difference between the measured $q_{src}^{max}$ and $q_1^{max}$: $q_3 = q_{src}^{max} - q_1^{max}$
*3 Estimated by (2) and the estimated $q_3$: $f_1 = \varepsilon \rho q_3^2 / s_3$
*4 Estimated by the measured $q_1^{max}$ and the relationship shown in Fig. 5(b).

$\varepsilon$, and the other parameters should be determined such that (3) is satisfied according to the required $q_1^{max}$ that determines the performance of the soft fingers.

*B. Finger Unit with Variable Surface Friction Mechanism*

This section describes the structure of the soft finger unit with a lubricant injection system for varying the contact surface friction. Fig. 9 shows a schematic of the developed finger unit. The finger unit consists of the main finger body, lubricant injection mechanism, slit-shaped surface texture, and lubricant tank. The components, except for the tank, were molded with a soft and deformable material (the main body of the finger and the lubricant injection mechanism, Dragon skin 10 Medium; surface texture, silicone sealant). The air chamber is embedded in the main finger body, and the finger motion is controlled by pressurization and relaxation of the airflow sent from the FCS mechanism through output tube 1. The finger surface is covered by a slit-shaped surface texture, as proposed in our previous study [9][10]. The texture provides a large surface friction, whereas the lubricant (anhydrous ethanol) can sufficiently reduce the friction to cause a slippage in the contact area [10]. The ethanol injected onto the surfaces of the finger and object was volatilized and disappeared after the operation. Little damage to the object caused by the injection, such as wrinkle formulation, occurs. The details are provided in [10].

In this study, the deformability and generable tip force of the soft finger were investigated. Fig. 10(a) shows the experiment setup used to measure the finger posture when the pressure of the input airflow was varied from 0 to 35 kPa. Red marks were placed on the side of the finger, and image processing was used to evaluate the posture of the finger. The coordinate frame, $\Sigma_A$, used for evaluating finger posture was set as shown in Fig. 10(a). Fig. 10(b) shows the relationship between the input pressure, $p_f$, and bending radius $r$. Fig. 11(a) illustrates the experiment setup used to measure the generable tip force, $f_{tip}$, and Fig. 11(b) shows the result. Next, the structure of the lubricant injection mechanism is described. The Venturi effect was used to operate the lubrication by drawing the lubricant from the lubricant tank with compressed air supplied from output tube 2. The pressure of the fluid is reduced with an increase in the velocity of the fluid flowing through the throttled section (orifice) of the flow path. Fig. 12 shows a schematic of the lubricant injection mechanism. As shown in Fig. 12(a), the mechanism consists of an airflow path with an orifice and two paths that branch off from the orifice and connect to the lubricant tank. The structure was molded into a single piece. Let $p_{in}$ and $p_{out}$ be the airflow pressure at the wide section before flowing to the orifice and the airflow pressure at the orifice, respectively; $v_{in}$ and $v_{out}$ be the airflow velocities at each section; and $s_{in}$ and $s_{out}$ be the cross-sectional areas of the flow paths at each section. Based on Bernoulli's principle, the difference in pressure between the two sections is given by

$$p_{in} - p_{out} = \rho(v_{out}^2 - v_{in}^2)/2 \quad (4)$$

When the volumetric airflow rate from the FCS mechanism ($q_2$) is applied through output tube 2, the continuity equation makes $q_2$ constant in the two sections. From $v_{in} = q_2/s_{in}$ and $v_{out} = q_2/s_{out}$, (4) can be expressed as follows:

$$p_{in} - p_{out} = \rho q_2^2 (1/s_{out}^2 - 1/s_{in}^2)/2. \quad (5)$$

From (5), $p_{in} > p_{out}$ is satisfied because $s_{out} < s_{in}$. Therefore, a negative pressure can be generated, regardless of $q_2$. From (5), $p_{out}$ decreases with an increase in $q_2$.

As discussed in Section III.A, to ensure a wide range of finger motions while stopping the lubricant injection, we proposed a device that injects lubricant only when $q_{src}$ exceeds the threshold of 118 L/min (in Prototype A), that is, in State C (see Fig. 3(c)). Fig. 12(b) shows the structure applied. It is assumed that the approaching direction of the robot hand is toward the direction of gravity, for example, to grasp an object placed on a table. The lubricant supplying ports of the injection mechanism are connected to the lubricant tank, mounted on the palm, through lubricant-supplying tubes. If the negative

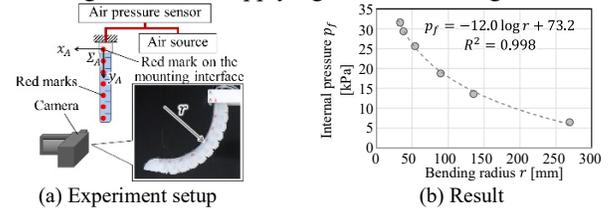
(a) Experiment setup  (b) Result
Fig. 10. Relationship between $p_f$ and $r$

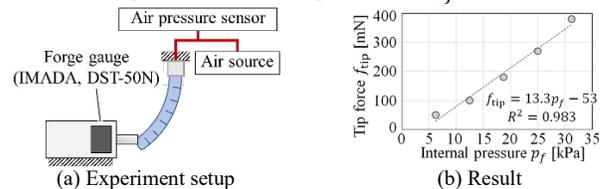
(a) Experiment setup  (b) Result
Fig. 11. Relationship between $p_f$ and $f_{tip}$

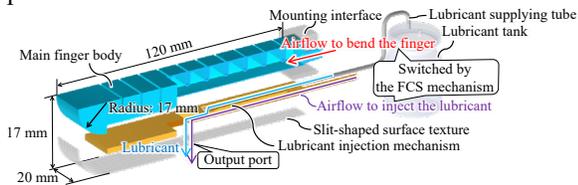
Fig. 9. Schematic view of finger unit.

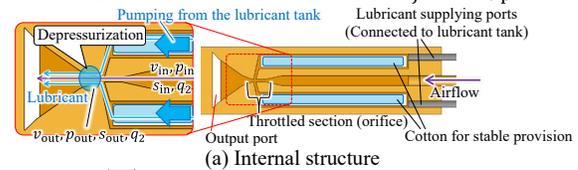
(a) Internal structure

(b) Device for controlling the relationship between injection and airflow
Fig. 12. Schematic view of the lubricant injection mechanism.

pressure, $p_{\text{out}}$, is generated by the airflow,
$$p_{\text{atm}} = \rho s_t h_l g + p_{\text{out}} \tag{6}$$
is obtained, where $p_{\text{atm}}$ is the magnitude of atmospheric pressure, $s_t$ is the cross-sectional area of the lubricant supplying tube, and $h_l$ is the height of the lubricant pumped into the tube, viewed in the direction of gravity from the top of the lubricant stored in the tank (Fig. 12(b)). Let $h_t$ be the maximum height of the tube. Then, if $h_l > h_t$, the lubricant flows into the orifice and is injected from the output port of the finger surface. From (5) and (6), the condition is
$$h_l = (p_{\text{atm}} - p_{\text{out}})/\rho s_t g$$
$$= \frac{1}{s_t}\left(\frac{p_{\text{atm}} + p_{\text{in}}}{\rho g} + \frac{g q_2}{2}\left(\frac{1}{s_{\text{in}}^2} + \frac{1}{s_{\text{out}}^2}\right)\right) > h_t. \tag{7}$$
From Bernoulli's principle, $p_{\text{in}}$ is given by
$$p_{\text{in}} = p_{\text{src}} - p_{\text{atm}} + \frac{\rho}{2}\left(\frac{q_{\text{src}}^2}{s_{\text{src}}^2} - \frac{(q_{\text{src}} - q_2)^2}{s_e^2} - \frac{q_2^2}{s_{\text{in}}^2}\right), \tag{8}$$
where $p_{\text{src}}$ is the pressure of air stored in an air source, and $s_{\text{src}}$ and $s_e$ are the cross-sectional areas of the outlets of the air source and exhaust port, respectively; hence, $p_{\text{in}}$ is determined by $q_2$, which is controllable through the airflow controller. $s_t$, $s_{\text{in}}$, $s_{\text{out}}$, and $h_t$ are the design parameters. By tuning them, the injection is activated only when $q_2 \geq 44$ or $q_{\text{src}} \geq 118$ L/min (i.e., in State C). For easy tuning, only $h_t$ was tuned in this study to prevent the injection in states A and B. In addition, the desired injection (i.e., activating only when $q_{\text{src}} \geq 118$ L/min) was achieved even if the approaching direction of the hand was oriented, as long as $h_l$ did not exceed the tuned $h_t$ when $q_{\text{src}} < 118$ L/min. Note that the desired activation of the lubricant injection can be achieved even with the robotic hand oriented in various directions if multiple tanks with different installation orientations are prepared, or if the mechanism used to maintain the tank orientation irrespective of the hand posture is provided.

## IV. EVALUATION OF INTEGRATED KEY MECHANISMS

In this section, we confirm the motion range of the finger and whether finger movement and lubricant injection can be operated independently. Fig. 13 shows $q_{\text{src}}$ for the finger motion control, the range for the lubricant injection, and the states of the FCS mechanism according to $q_{\text{src}}$. The amount of lubricant required to create a slippery surface can be set to a constant value. Meanwhile, the motion of the robot finger must be controlled by controlling the flow rate in small steps. Hence, we adopted a method by which the airflow controller (Kofloc, 8550MC) within a range of 0−50 L/min was used to control the finger motion, whereas the valve of the controller was fully opened for lubricant injection. An air source (Prostyle, PCR3010) was used, and the flow rate of 150 L/min when the valve was fully opened was sufficiently high to inject the lubricant while maintaining the finger posture (over 118 L/min). Note that, to ensure robustness against environmental changes such as changes in $h_t$ (the height of the lubricant in the tank) owing to the injection and tilt of the hand, as well as a change in $p_{\text{atm}}$ based on the weather conditions, the lubricant injection was applied when the valve was fully opened such that the airflow rate for the injection was isolated from the range for the finger motion control.

### A. Finger Motion

This section evaluates the finger motion through the airflow from the FCS mechanism. The experiment setup is shown in Fig. 14. In this study, the robot hand was composed of two developed fingers, and the fingers were controlled simultaneously through the same airflow. Because the two fingers were identical, only one finger was evaluated in this setup. The tube fixer was installed on the palm of the hand to tune $h_t$ and control the value of the airflow that activates the lubricant injection. Through trial and error, $h_t$ was set to 55 mm. The finger posture was obtained through image processing, as shown in Fig. 10, when the input airflow ($q_{\text{src}}$) was varied from 0 to 50 L/min. We observed that the lubricant injection did not operate within an airflow range of 0−50 L/min. The results are presented in Fig. 15. We confirmed that the finger bent to such an extent that it could grasp and manipulate objects by controlling the input airflow rate.

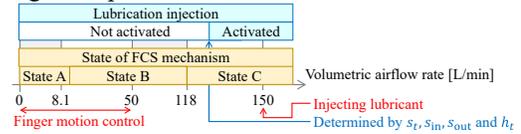

**Fig. 13.** Range of the volumetric airflow rate of the air source ($q_{\text{src}}$) for the finger motion control, range for the lubricant injection, and states of the FCS mechanism according to $q_{\text{src}}$.

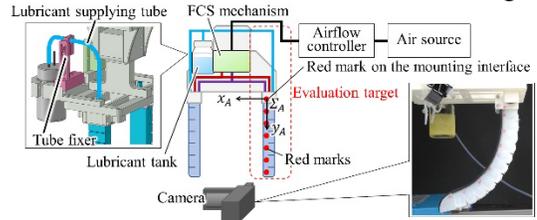

**Fig. 14.** Experiment setups for observing the finger motion.

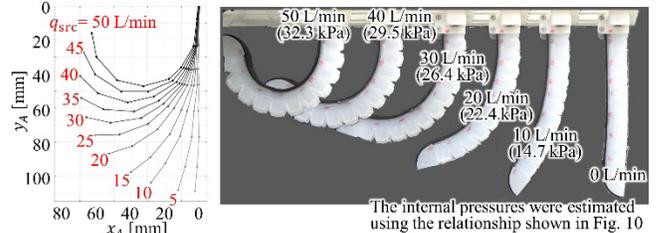

**Fig. 15.** Finger motion when changing input airflow rate.

### B. Lubricant Injection

This section confirms whether the lubricant injection is activated while maintaining the finger posture when the valve of the controller is fully opened ($q_{\text{src}} = 150$ L/min). The experiment setup is the same as that shown in Fig. 14. First, input airflow rates of 10, 20, 30, 40, and 50 L/min were supplied. The rate was then switched to 150 L/min by opening the valve to inject the lubricant. The postures before and after injection were evaluated using image processing. The displacements of the eight red marks placed on the side of the finger before and after injection were evaluated. Let $p_i \in \mathbb{R}^2$ be the position of the $i$th mark, and their mean displacement, $p_{\text{disp}}$, before and after the injection are derived as

$$p_{\text{disp}} = \Sigma_{i=1}^{8} \|\boldsymbol{p}_i^{after} - \boldsymbol{p}_i^{before}\|/8. \quad (9)$$

Fig. 16 shows the results for each airflow rate. The displacement was sufficiently small considering the difficulty of controlling the soft robotic finger to within an error of 1 mm. The finger posture was maintained before and after injection, regardless of the airflow rate. The injection function was not operated when the airflow rate was less than 50 L/min and was operated at a rate of 150 L/min.

## V. ROBOTIC HAND SYSTEM

This section evaluates the total robotic hand system shown in Fig. 14, which integrates the FCS mechanism and pneumatic soft fingers with variable friction mechanisms.

### A. Grasping Test

The developed hand attached to the automatic positioning stage grasps an object (Fig. 17(a)), placed on a table. Most of the objects were grasped (Fig. 17(b)), whereas a small binder clip could not be grasped because of the gap between the table surface and fingertip, caused by the arced trajectory of the fingertip (see Fig. 17 (c)). To grasp small objects, a grasping strategy utilizing finger posture control by controlling the input airflow rate was proposed. The fingers were closed such that their opening width was slightly wider than the object width, and the robotic hand grasped the object when the fingertips approached the table surface to a distance sufficiently close to touch it. This strategy eliminates the gaps caused by the arching of the finger movements and allows the grasping of small objects (see Fig. 17(d)). The results demonstrate the efficacy of the developed hand in terms of object grasping (see also the attached video clip). From the results shown in Figs. 11 and 15, the generable tip force applied to an object was estimated to be 380 mN with an airflow of 50 L/min (32.3 kPa). If assuming a friction coefficient of 2.0, the robotic hand would obtain a payload of 1.5 N (0.38 N × 2 fingers × 2.0). We also demonstrated the payload experimentally; in addition, the width of the graspable object was under 73 mm.

### B. Placement Accuracy

Here, we evaluated the effect of varying the finger surface friction on the accuracy of the placement. The activated slippage makes it possible to release an object without changing the finger posture [10]. This eliminates the uncertainties caused by the opening motion of the fingers and ensures an accurate placement. We evaluated the accuracy of the results when placing an object with and without lubrication, that is, the change in surface friction. The experiment setup is shown in Fig. 18. The developed hand grasps the rectangular target object placed on a table and lifts it up by 20 mm. When the lubrication was not activated, the hand moved downward by 20 mm and was then opened. When using lubrication, the lubricant was injected after lifting. The slippage at the contact area between the object and fingers achieves object placement. The experiments were conducted 10 times for each condition. To eliminate the effect of the error during the grasping process, the displacements of the object position and posture during and after grasping were evaluated using the camera image (see Fig. 18). The translational displacement of the bottom face of the object, $\Delta d$, and the rotational displacement (around the $z_B$ axis), $\Delta\theta$, from the position and posture during grasping were evaluated. The results for $\Delta d$ and $\Delta\theta$ are presented in Fig. 19. Both the $\Delta d$ and $\Delta\theta$ with lubrication were smaller than those without a lubricant. This is thought to be because the finger works as a guide when the object slides on the surface of the finger (see also the video clip).

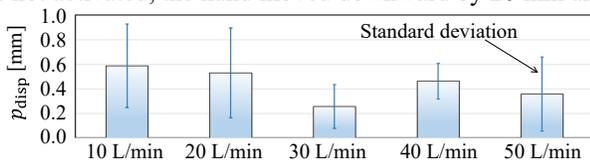

**Fig. 16.** Difference in finger posture before and after fully opening the valve to inject the lubricant.

### C. Manipulation

Using a pivot manipulation, we verified that slippage at the contact surface, which is difficult to activate on a soft finger surface, facilitates manipulation. The target manipulation was to rotate a rectangular silicone object on the table around the contact edge between it and the table surface. The experiment setup was the same as that shown in Fig. 18. The procedure is illustrated in Fig. 20. We compared the results with and without lubricant injection. The results are shown in Fig. 20. In the manipulation without lubrication, attempts to rotate the object the manipulation to fail. This is mainly due to the large friction between the fingertip and the object. By contrast, manipulation with lubrication was successful owing to rotational slippage at the contact area between the fingertip and the object caused by the lubrication.

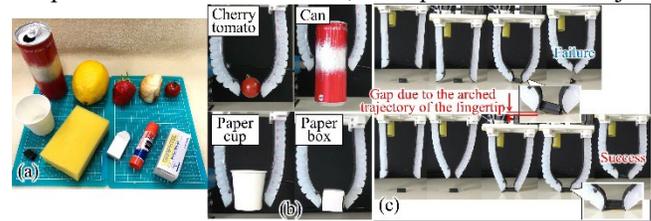

**Fig. 17.** Results of the grasping tests: (a) Target objects, (b) representative results, (c) test of the small binder clip (top: w/o a grasping strategy and bottom: w/ a grasping strategy).

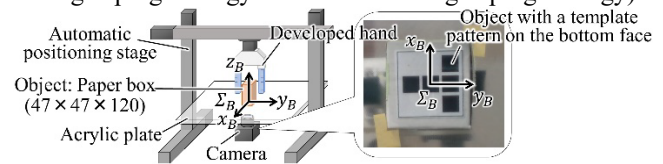

**Fig. 18.** Experiment setup to evaluate the placing ability.

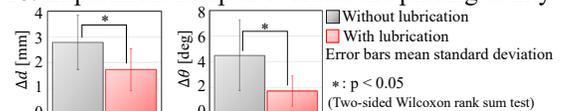

**Fig. 19.** Displacements of positions and posture during grasping and after placement with and without lubricant injection (the values are absolute values).

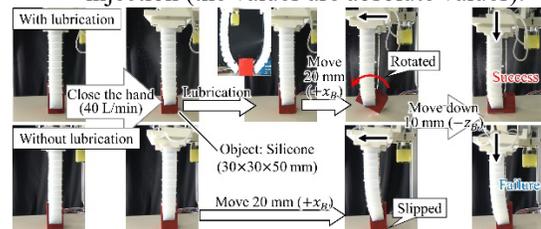

**Fig. 20.** Images of pivoting with and without lubrication resulted in slippage between the object and the table, causing

## VI. CONCLUSION

A soft robotic hand equipped with a novel flow switching system was proposed that can switch the airflow path using a single airflow control, i.e., an FCS mechanism. The presented design and experiment evaluations demonstrated that all functional requirements were met. The output airflow from the switching system is provided to the pneumatically driven soft finger of the hand and the lubricant injection system to reduce the high friction of the finger surface. Employing the system, the developed hand achieved the switching of two completely different functions: finger motion control and lubricant injection operation through 1-DOF actuation. In the lubricant injection mechanism, positive air pressure was converted into negative air pressure using the Venturi effect to reduce the friction of the finger surface. Several evaluations demonstrated the effectiveness and performance of the key mechanisms, i.e., the FCS mechanism and pneumatically driven soft finger unit with a friction variable mechanism. The grasping, placing, and manipulation tests also demonstrated that the developed hand system worked effectively, and the slippage activated by the lubrication facilitated the release and pivot manipulation. Our future study will involve an optimization of the design parameters of the FCS mechanism, such as the dimensions of the switching lever, and more general grasping and motion planning with the developed robotic hand when utilizing lubrication.